# INTERPRETABLE SELF-ATTENTION TEMPORAL REASONING FOR DRIVING BEHAVIOR UNDERSTANDING

*Yi-Chieh Liu[1,*], Yung-An Hsieh[2,*], Min-Hung Chen[2], C.-H. Huck Yang[2], J. Tegner[3], Y.-C. James Tsai[1,2]*

[1]School of Civil and Environmental Engineering; [2] School of Electrical and Computer Engineering
Georgia Institute of Technology, Atlanta, GA, USA
[3]Living Systems Laboratory, KAUST, KSA

## ABSTRACT

Performing driving behaviors based on causal reasoning is essential to ensure driving safety. In this work, we investigated how state-of-the-art 3D Convolutional Neural Networks (CNNs) perform on classifying driving behaviors based on causal reasoning. We proposed a perturbation-based visual explanation method to inspect the models' performance visually. By examining the video attention saliency, we found that existing models could not precisely capture the causes (e.g., traffic light) of the specific action (e.g., stopping). Therefore, the Temporal Reasoning Block (TRB) was proposed and introduced to the models. With the TRB models, we achieved the accuracy of **86.3%**, which outperform the state-of-the-art 3D CNNs from previous works. The attention saliency also demonstrated that TRB helped models focus on the causes more precisely. With both numerical and visual evaluations, we concluded that our proposed TRB models were able to provide accurate driving behavior prediction by learning the causal reasoning of the behaviors.

## 1. INTRODUCTION

Human drivers are required to perform traffic scenes reasoning and act correspondingly to ensure driving safety. Therefore, to design a robust self-driving system, two important factors need to be considered as shown in Fig.1. First, a reasoning model is needed to predict actions based on the reasoning that human drivers perform. Second, a visualization of video attention saliency is required to check and further improve the models on predicting the behaviors based on the correct reasons.

Various driving scene datasets have been introduced [1, 2, 3, 4, 5] to accelerate the pace of self-driving vehicles development. Most of the datasets contain annotations of traffic scenes (e.g. segmentation of pavements [1]) or visual-based control (e.g. steering angles [5]). On the other hand, Ramanish *et al.* [4] proposed the Honda Research Institute Driving Dataset (HDD), which contains the annotations of higher-level driving behavior classes. A 4-layer annotation was defined in the HDD to describe the driving behaviors. To explore the causal reasoning on self-driving vehicles, the Stimulus-driven Action and Cause layers were utilized in this work. A detailed description of the data was shown in Section 4.1.

With the ability to consider both spatial and temporal relationships, video recognition models have proven their effectiveness on action recognition tasks [6, 7, 8]. Video recognition models also benefit the development of algorithms for self-driving vehicles [9, 10]. By providing video inputs, the models can predict control signals, such as steering angles, for self-driving vehicles. In this work, video recognition models were used as the reasoning model of driving behaviors, in which the behaviors based on causal reasoning (e.g. stop for traffic light) were classified. Video recognition models can be categorized into Convolutional Recurrent Neural Networks (CRNNs) [11] and 3D Convolutional Neural Networks (CNNs) [12]. While [4] proposed a CRNN as a reasoning model of driving behaviors, the performance of 3D CNNs on those higher-level behaviors has not been explored. To further improve the models on reasoning driving behaviors, we introduced a **Temporal Reasoning Block (TRB)** to enhance the model understanding of the causes of driving behaviors. The experiment showed that TRB could help the models capture spatial-temporal features and global dependency within videos using a self-attention mechanism.

[*]Equal contribution.

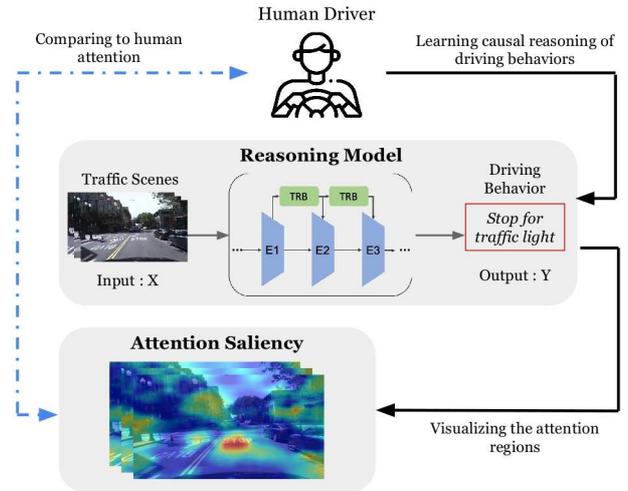

**Fig. 1**: Designing a robust self-driving system includes two important factors. **Reasoning Model** aims at predicting driving behaviors based on human causal reasoning. Our proposed TRB demonstrated its effectiveness in improving the state-of-the-art models. **Attention Saliency** helps examine and optimize the system to align with the mechanism of how humans interact with complex environments.

The visual traffic scenes usually contain complex information, such that different observed objects can serve as the cause of a certain action in different scenarios. For example, both traffic lights and pedestrians occur in two scenarios in which the car is stopping. In one scenario, the stopping is caused by the traffic light, while the other is caused by the pedestrians crossing the road. To inspect whether the model is paying attention to the actual cause of action, a visual explanation of CNNs is required. The visual explanation aims at understanding where the models are paying attention to. In other words, it generates the attention saliency of the input data. To inspect the models for self-driving vehicles, we introduced a spatial-temporal explanation method. **Our contributions include:**

- The investigation of state-of-the-art 3D CNNs on the recognition of driving behaviors based on causal reasoning.
- The introduction of the Temporal Reasoning Block (TRB) for improving the state-of-the-art models on classifying reasoning-based driving behaviors.
- The proposition of a perturbation-based visual explanation method for spatial-temporal models, which enables the inspection of self-driving models.

## 2. RELATED WORK

### 2.1. Self-Driving Behavior Recognition

As self-driving technology demonstrated incredible performance in both urban and off-road scenarios [13], the reasoning of self-driving behavior became a needed research problem. ALVINN [14] proposed a shallow neural network to classify actions from images. Researches have also suggested the potential of deep neural networks for tightly coupling the perception and control in simple scenarios [9]. Robust action-perception mapping models were also been explored to recognize the complex visual representation in urban environments [10]. However, these prior efforts formulate the behavior as a goal-oriented task, which is not sufficient to learn how humans drive and interact with traffic scenes. In our work, we explore the cause-aware features by our proposed framework using the HDD [4]. Driving behavior understanding could also be performed by video recognition approaches. One of the approaches is combining CNNs and Recurrent Neural Networks (RNNs) to model the temporal pattern of visual representation [12]. 3D convolutions (C3D) [15] and its improvements [7] also shown great success in space-time features. Also, the 2D residual architecture is also extended into the 3D residual CNNs (3DResNet) [16].

### 2.2. Attention Models

Recently, attention mechanisms have become a reliable method to capture global dependencies [17, 18]. In particular, self-attention [19] represents the importance of different positions in a sequence. The mechanism had been applied to actions recognition tasks in video as a non-local operation [20]. Yet, the potential of self-attention have not been explored on the reasoning tasks of driving behaviors.

### 2.3. Visual Explanation of CNNs

Different methods have been proposed to visually explain where CNNs are paying attention to the input images. Class Class Activation Mapping (CAM) [21] and Grad-CAM [22] generate visual explanation by linear combining activations and class-specific weights. Backpropagation methods were used in DeConvNet [23]. A problem of the above methods is that accessing intermediate layers [21, 22, 24] and/or architectural modification [23] are required. On the other hand, several methods perform visual explanation by perturbing the input images. The dropping of classification score was observed when occluding a small patch on the image in [23], while [25] occluded the image with segmented super-pixels. In [26], the perturbed regions on the image were found by solving an optimization problem. This method can be used on any kind of the model since it is based on explicitly perturbing the images. Therefore, we adapted this method to the visual explanation of self-driving models.

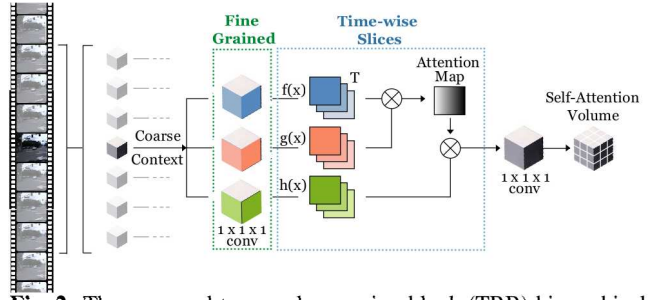

**Fig. 2**: The proposed temporal reasoning block (TRB) hierarchical structure. The $\otimes$ denotes matrix multiplication. We applied softmax operation on each row to generate the attention map for each frame and stack them back to spatial-time volume to acquire the temporal-aware self-attention features.

## 3. METHODOLOGY

### 3.1. Temporal Reasoning Block

Most of the non-local based video understanding [20] models optimize the performance with spatial-temporal dependency. However, for the reasoning-oriented dataset, spatial features and temporal integrity should be more weighted to explore the causes of behavior. Inspired by the non-local operation [20], we proposed **temporal reasoning block (TRB)**, a hierarchical self-attention module on different dimensions to reason the driving behavior, as shown in **Fig. 2**. In our proposed method, the TRB consists of two layers which focus on spatial-temporal and spatial domain separately and takes input from previous state-of-art feature extractor. The input $x \in \mathbb{R}^{\mathbb{C} \times \mathbb{T} \times \mathbb{N}}$ could be seen as coarse-grained video context. In the first layer of TRB, we used 3D convolution kernel to extract fine-grained spatial-temporal feature $h(x) \in \mathbb{R}^{(\mathbb{C} \times \mathbb{T}) \times \mathbb{N}}$ from $x$ to keep the temporal continuity in the second layer, we sliced $x$ temporal-wise to $x_s \in \mathbb{R}^{\mathbb{C} \times \mathbb{N}}$ and transformed each $x_s$ into two feature spaces $f$ and $g$ to attain spatial attention, where $f(x)=W_f x, g(x)=W_g x$,

$$\alpha_{j,i} = \frac{exp(s_{ij})}{\sum_{i=1}^{N} exp(s_{ij})}, s_{ij} = f(x_i)^T g(x_j) \quad (1)$$

$\alpha_{j,i}$ represents the attention score of the $i^{th}$ location when interacting with the $i^{th}$ region. $C$ is the number of channels and $N$ is the number of feature locations of the acquired feature map, where $N = W \times H$, $W$ and $H$ are the size of the feature map. The output $o$ represents the stacked attention map along with time domain, where,

$$o_j = \sum_{i=1}^{N} \alpha_{j,i} h(x), \ h(x) = W_h x \quad (2)$$

$$O_v = Stack\{o_t\}, t = 1 \ to \ T \quad (3)$$

In above formula, $O_v \in \mathbb{R}^{\mathbb{C} \times \mathbb{T} \times \mathbb{W} \times \mathbb{H}}$. For best describing temporal characteristic and memory efficiency, we choose $\bar{C} = C/8$ in our all experiments. We further scaled the attention volume and added it back to the input $x$. The final output could be represented as,

$$Y_i = \gamma O_i + x_i \quad (4)$$

where $\gamma$ is a learnable scalar parameter.

## 3.2. Visual Explanation for Self-Driving Models

In [26], the visual explanation was done by finding the regions to perturb the original image which makes the classifier model, $f_c$, to produce a minimal score on the target class. This is done by defining a mask $m : \Lambda \to [0, 1]$, where $\Lambda = \{1, ..., H\} \times \{1, ..., W\}$. The perturbed image can be defined as:

$$[\Phi\{x_0; m\}](u) = m(u)x_0(u) + (1 - m(u))x_{p(u)} \quad (5)$$

where $u \in \Lambda$ is each pixel location and $x_p$ is generated by performing perturbation on the whole original image $x_0$. In other words, the more $m(u)$ is closer to 0, the more severe the pixel $u$ will be perturbed. Different kinds of perturbation operations can be used, such as add noise or blurring the image. To find the optimal mask $m^*$ that minimizes the classification score $f_c(\Phi\{x_0; m\})$, a objective function is defined as:

$$\min_{m \in [0,1]^\Lambda} f_c(\Phi\{x_0; m\}) + \lambda_1 \|1 - m\|_1 + \lambda_2 \sum_{u \in \Lambda} \|\nabla m(u)\|_\beta^\beta \quad (6)$$

where the L1 term aims at making the mask as small as possible and the total variation (TV) regularization aims at smoothing the mask. To prevent the mask from over-fitting the network, a lower resolution mask is solved during the optimization process. The mask is up-sampled to the image size when performing the perturbation.

To visually inspect self-driving models, the perturbation-based method needs to be extended to perform the spatial-temporal explanation. A three dimensional mask is now used, such that $m : (\Lambda, T) \to [0, 1]$, where $T$ is the size of the additional temporal dimension. To consider the relationship between each frame of the input video, a temporal TV regularization term is introduced:

$$\sum_{t \in T} \lambda_t \|\nabla m(:, t)\|_\beta^\beta \quad (7)$$

where $\lambda_t$ is the regularization coefficient. By adding the third dimension in eq. (6) and the temporal tv regularization, the objective function can now be written as:

$$\min_{m \in [0,1]^{(\Lambda,T)}} f_c(\Phi\{x_0; m\}) + \lambda_1 \|1 - m\|_1 + \sum_{t \in T} \left( \lambda_s \sum_{u \in (\Lambda, t)} \|\nabla m(u, t)\|_\beta^\beta + \lambda_t \|\nabla m(:, t)\|_\beta^\beta \right) \quad (8)$$

where $\lambda_s$ is the coefficient of the spatial tv regularization. Similarly, a lower resolution mask in both spatial and temporal dimensions is used to prevent the mask from over-fitting the network.

## 4. EXPERIMENTAL ENVIRONMENT

Two experiments were conducted to evaluate and inspect the proposed TRB models on the reasoning of driving behaviors. A baseline CRNN and the 3D CNNs, both with and without the proposed TRB, were trained and evaluated. Then the proposed perturbation-based explanation method was applied to generate the video attention saliency, which was used to inspect where the models were paying attention to.

### 4.1. Dataset Preparation

The annotation layers of Stimulus-driven Action and Cause in the HDD were utilized. The Stimulus-driven Action contains *stop* and *deviate*, while Cause contains the reasons of these two actions. To focus the experiment on the reasoning for driving behaviors, we simplified the task to consider a single action class. Stopping is a fundamental but critical behavior to ensure driving safety. Therefore, the stop action was selected, with the causes of traffic light (stop4light), pedestrian (stop4ped), stop sign (stop4sign), and congestion (stop4cong).

Video clips were prepared as the input of the models. Each clip contains 20 continuous frames with 4 frame-per-second, and each frame image has a size of 360 by 360. Each clip also has a target label from the 4 classes. The numbers of video clips for each class were shown in **Table 1**.

| Data splits | stop4light | stop4ped | stop4sign | stop4cong |
|---|---|---|---|---|
| Train | 100 | 45 | 170 | 170 |
| Validation | 10 | 6 | 20 | 20 |
| Test | 13 | 10 | 30 | 30 |

**Table 1**: The number of video clips for each class.

### 4.2. Model Implementation

**Baseline CRNN**: The *CNN conv* model in [4] was set up as the baseline in the experiment. The InceptionResnet-V2 model was served as the CNN encoder. The encoded features were flattened and inputted to a Long-Short Term Memory (LSTM) network. The output of the model was generated from the last LSTM hidden state by connecting a fully connected (FC) layer with the size of four.

**3D models**: All 3D CNNs were implemented base on the *Conv3d* layer in Pytorch. The C3D [15], I3D [7], and 3DResnet [16] models were implemented and fine-tuned on the prepared HDD. The Resnet34 [27] architecture was used in the 3DResnet model. We observed that adding TRB in shallow layers of the models did not improve the performance. Therefore, we added four TRBs in the dur layers of each model.

**Training**: The models were trained on a single GeForce RTX 2080Ti GPU. Cross entropy loss was used as the objective function for the recognition task. Stochastic gradient descent (SGD) was used as the optimizer, with a learning rate of 0.0001, a momentum of 0.9, and a weight decay of 0.0001. The models were trained with 150 epochs and the batch size set to 4. Early stopping was applied to select the weights of the model that were used in the evaluation stage. To deal with the imbalanced training samples for different classes, oversampling of data was used during the training process.

**Evaluation**: Different from [4], which performed per-frame evaluation on testing videos and calculated the average precision(AP), we measured the model performances based on the video recognition tasks. Video clips, where each contains a single target label, were used for evaluation. The accuracy of classifying the label for each video clip in the test set was computed for each model.

### 4.3. Visual Explanation

The visual explanation was generated using Adam optimizer to minimize eq. (8). The following parameters were used in the experiment: learning rate = 0.01, iteration = 500, $\lambda_1 = 0.001$, $\lambda_s = 0.2$, $\lambda_t = 0.1$, and $\beta = 3$. The image was perturbed by combining both Gaussian and median blur. The perturbation mask $m$ was set to the size of $(28 \times 28 \times 10)$. The mask was up-sampled to the size of the video clip and served as an attention saliency $M$, which larger intensity indicated more attention.

To quantitatively examine how much the models pay attention to a certain object, we defined an attention score, $S_{Atten}$, for each

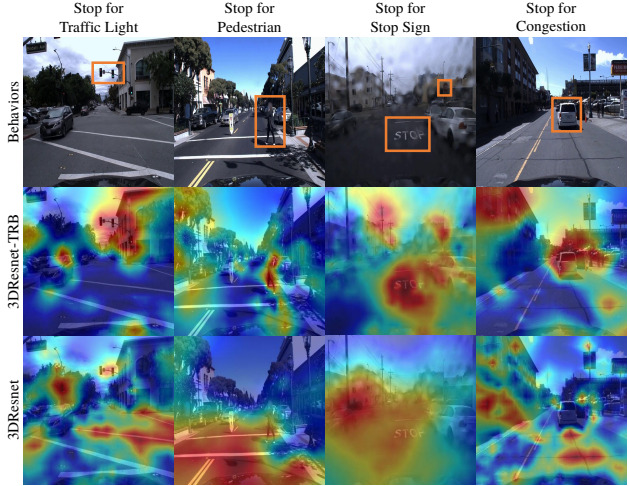

**Fig. 3**: Attention saliency of each driving behavior class. The first row showed the frames with the causes labeled by bounding boxes. The second and third rows showed the attention saliency of 3DResnet-TRB and 3DResnet, respectively.

object $o$ in a single frame $I_t$ as:

$$S_{Atten} = \sum_{u \in I_t} s_u, \text{ where } s_u = \begin{cases} M(u,t) & d(u,c_o) \leq r_o \\ \frac{M(u,t)}{d(u,c_o)} & d(u,c_o) > r_o \end{cases} \quad (9)$$

In eq. (9), $M(u,t)$ is the value of the attention saliency at a pixel location $u$ in time $t$. $d(u, c_o)$ is the distance between $u$ and the center of the object $o_c$, while $r_o$ is the radius of the circle that encloses the object. To compensate the different sizes of the objects, we normalized the attention score when comparing different objects.

## 5. EXPERIMENTAL RESULTS

### 5.1. Driving Behavior Recognition

The evaluation results of each model on classifying the driving behaviors were presented in **Table 2**. Among the 3D CNNs, both I3D and 3DResnet showed better performance than the baseline CRNN. This indicated that the state-of-the-art 3D CNNs were able to more precisely classify the reasoning-based driving behaviors. With the highest accuracy among the models without TRB, 3DResnet also demonstrated the effectiveness of the residual architecture on this task. As shown in the table, the proposed TRB largely improved the performance of all models. This indicated that the self-attention mechanism in TRB effectively helped the models to capture the global dependency within the videos. The result also showed that the TRB can be flexibly applied to different models of driving behavior recognition to provide improvement.

### 5.2. Attention Saliency of Driving Behaviors

Examples of generated attention saliency using the proposed visual explanation method were shown in **Fig. 3**. The frame images were shown in the first row, with the causes of the action (e.g. stop signs) being labeled by bounding boxes. The attention saliency was presented as heatmaps in the second and third rows. As shown in the figure, adding the proposed TRB to 3DResnet made the model capture the causes more precisely. For each of the class, 3DResnet-TRB paid attention to most of the regions around the bounding boxes. This demonstrated the effectiveness of TRB on the reasoning of driving behaviors.

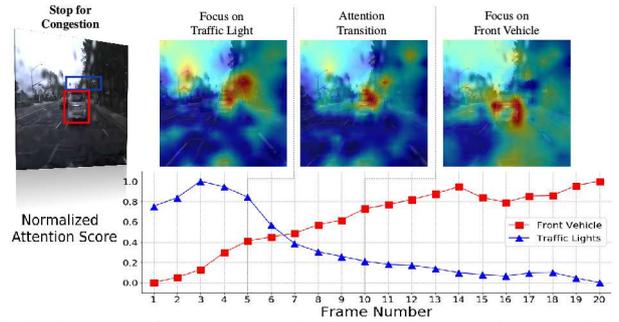

**Fig. 4**: The attention scores of the frond vehicle (red) and traffic lights (blue) in a stop for congestion behavior occurred in a rainy scenario. As the vehicle approached the front vehicle, the attention transition from traffic lights to the front vehicle was observed.

| Model | Accuracy | Model | Accuracy |
|---|---|---|---|
| CRNN | 73.49% | CRNN-TRB | 78.31% |
| C3D | 60.71% | C3D-TRB | 69.88% |
| I3D | 77.11% | I3D-TRB | 83.13% |
| 3DResnet | 83.56% | 3DResnet-TRB | 86.30% |

**Table 2**: The recognition results of driving behaviors based on causal reasoning.

**Fig. 4** presented a quantitative evaluation of attention saliency generated from 3DResnet-TRB. A stop for congestion behavior in the rainy scenario was showed. The attention score of the front vehicle and the average score of multiple traffic lights (enclosed by the blue bounding box) were computed. A reasonable attention transition can be observed from both the attention scores and attention saliency. When the vehicle approached the front vehicle, the model gradually shifted its attention from the traffic lights to the front vehicle, which was the cause of the stopping behavior. This example indicated that the proposed TRB helped the models pay attention to reasonable regions when predicting driving behaviors.

## 6. CONCLUSIONS

In this work, we proposed the **Temporal Reasoning Block (TRB)** to improve the performance of video recognition models on reasoning driving behaviors. The results showed that TRB can largely improve the state-of-the-art models, both CRNN and 3D CNNs, on classifying driving behaviors. By adding TRB to the 3DResnet, we achieved the highest accuracy of 86.3%. A perturbation-based visual explanation method was also proposed to generate video attention saliency. The examination of the attention saliency demonstrated that 3DResnet-TRB was able to focus on reasonable objects when classifying driving behaviors. With both numerical and visual evaluations, we concluded that our proposed TRB models were able to provide accurate driving behavior prediction by learning the causal reasoning of the behaviors.